\documentclass{article}

\usepackage{myarxiv}

\usepackage[utf8]{inputenc} % allow utf-8 input
\usepackage[T1]{fontenc}    % use 8-bit T1 fonts
\usepackage{hyperref}       % hyperlinks
\usepackage{url}            % simple URL typesetting
\usepackage{booktabs}       % professional-quality tables
\usepackage{amsfonts}       % blackboard math symbols
\usepackage{nicefrac}       % compact symbols for 1/2, etc.
\usepackage{microtype}      % microtypography
\usepackage{lipsum}		% Can be removed after putting your text content
\usepackage{graphicx}
\usepackage[numbers]{natbib}
\usepackage{doi}
\usepackage{latexsym}
\usepackage{stmaryrd}
\usepackage{amsmath,amssymb,amsfonts}
\usepackage{multicol,lipsum}
\usepackage{subcaption}
\usepackage{amsmath,amssymb}

\usepackage{threeparttable}

\usepackage{enumitem}

\usepackage{multirow}

\title{Syntactic Fusion: Enhancing Aspect-Level Sentiment Analysis Through Multi-Tree Graph Integration}

\author{ 
Jane Sunny\\
University of Charleston\\
\And
Tom Padraig\\
University of Charleston\\
\And
Roggie Terry\\
University of Charleston\\
\And
Woods Ali\\
University of Charleston\\
}

\renewcommand{\shorttitle}{Syntactic Fusion: Enhancing Aspect-Level Sentiment Analysis Through Multi-Tree Graph Integration}

\newcommand{\methodname}{SynthFusion}

\hypersetup{
pdftitle={aaa},
pdfsubject={aaa},
pdfauthor={aaa},
pdfkeywords={aaa},
}

\begin{document}
\maketitle

\begin{abstract}
Recent progress in aspect-level sentiment classification has been propelled by the incorporation of graph neural networks (GNNs) leveraging syntactic structures, particularly dependency trees. Nevertheless, the performance of these models is often hampered by the innate inaccuracies of parsing algorithms. To mitigate this challenge, we introduce SynthFusion, an innovative graph ensemble method that amalgamates predictions from multiple parsers. This strategy blends diverse dependency relations prior to the application of GNNs, enhancing robustness against parsing errors while avoiding extra computational burdens. SynthFusion circumvents the pitfalls of overparameterization and diminishes the risk of overfitting, prevalent in models with stacked GNN layers, by optimizing graph connectivity. Our empirical evaluations on the SemEval14 and Twitter14 datasets affirm that SynthFusion not only outshines models reliant on single dependency trees but also eclipses alternative ensemble techniques, achieving this without an escalation in model complexity.
\end{abstract}


\section{Introduction}

Aspect-level sentiment classification is a nuanced task within sentiment analysis, focused on discerning the sentiment (positive, negative, or neutral) towards specific aspects in textual content. For instance, in the sentence ``\textit{The exterior, unlike the food, is unwelcoming.}'', the sentiments towards ``exterior'' and ``food'' are negative and positive respectively \citep{nazir2020issues,FeiLasuieNIPS22,kiritchenkodetecting,dong2014adaptive}.
This task is particularly useful in parsing online reviews or aiding purchasing decisions on e-commerce platforms.

Recent studies highlight the effectiveness of syntactic structures, such as dependency trees, in capturing complex syntactic relationships often overlooked in superficial text analysis~\cite{zhang2018graph}. \nocite{FeiDiaREIJCAI22,Wu0RJL21,xiang2022semantic,FeiZRJ20,tang2015effective,Wu0LZLTJ22,ma2017interactive,fei-etal-2020-cross,chen2017recurrent,FeiTransiAAAI21,zhang2020convolution,WuFRLLJ21,chen2020inducing,FeiMatchStruICML22,huang2020syntax,FeiGraphSynAAAI21,9634849,hou2019selective,wwwLRZJ22}
Techniques employing graph neural networks (GNN) over dependency trees for aspect-level sentiment classification have shown promising results, linking aspect terms with relevant opinions for robust sentiment analysis~\cite{huang2019syntax,zhang2019aspect,sun2019aspect,wang2020relational}. However, these methods are not immune to the errors in parsing algorithms~\cite{wang2020relational}. For example, a misparsed sentence could lead to incorrect sentiment association for the term ``food''.

Acknowledging the limitations of current dependency parsers, especially in scenarios diverging from their training domain, we propose \methodname{}, a graph ensemble technique to mitigate parsing inaccuracies. This approach is grounded in the observation that different parsers, particularly those with distinct biases, often make unique errors \citep{Li00WZTJL22,fei2020enriching,dong2014adaptive,tang2015effective,fei2020boundaries}. By combining dependency trees from various parsers before implementing GNNs, our technique allows the model to concurrently assess multiple parsing hypotheses, thereby enhancing its decision-making process.

\methodname{} fuses edges from all provided dependency trees to create a comprehensive ensemble graph for GNN application. This exposes the model to diverse graph structures, empowering it to discern and prioritize more relevant edges for the given task. To maintain the syntactic relationships from the original trees, we categorize edges into parent-to-children and children-to-parent types, applying relational graph attention networks (RGAT)~\cite{busbridge2019relational} to the ensemble graph.

Our approach presents several advantages. Firstly, \methodname{}'s integration of multiple dependency trees allows GNNs to navigate through various parsing scenarios, enhancing the model's resilience to parsing errors. Moreover, this is achieved without incurring additional computational costs, as the GNNs operate on a singular graph. Additionally, by introducing more edges and reducing graph diameter, \methodname{} prevents GNNs from overfitting and limits the need for over-parameterization. This facilitates the use of shallower GNN models, thereby circumventing the common issue of over-smoothing in deeper networks~\cite{li2018deeper}.

In summary, our work contributes the following:
\begin{itemize}
    \item Introducing \methodname{}, an innovative technique that amalgamates dependency trees from diverse parsers to bolster model resilience against parsing errors. This ensemble graph approach allows the model to learn from noisy data and discern the most relevant edges without extra computational demands.
    
    \item Maintaining syntactic dependency integrity from the original trees by uniquely parameterizing edges, thereby enhancing the RGAT model's performance on the ensemble graph.
    
    \item Demonstrating that our \methodname{} RGAT model outshines recent leading models on three benchmark datasets (Laptop and Restaurant reviews from SemEval 2014 and the ACL 14 Twitter dataset), surpassing both single-parse models and other ensemble strategies.
\end{itemize}

\section{Related Work}

The field of aspect-level sentiment classification has witnessed significant advancements, with many studies emphasizing the application of attention mechanisms to sequential models. These attention mechanisms include co-attention, self-attention, and hierarchical attention, which have been effectively incorporated into recurrent neural networks (RNNs) for enhanced sentiment analysis~\cite{tang2015effective,liu2017attention,wang2018target,fan2018multi,chen2017recurrent,zheng2018left,wang2018learning,li2018hierarchical,li2018transformation}. These approaches leverage the sequential nature of language to better capture the nuances of sentiment associated with specific aspects in text.

Additionally, the advent of pretrained transformer language models, notably BERT~\cite{devlin2018bert}, has introduced a new dimension to this task. Unlike traditional models that rely on engineered features or complex architectures, BERT and similar models utilize deep bidirectional representations learned from vast amounts of unlabeled text. This has enabled more direct and effective handling of word sequences in sentiment classification, as demonstrated in various studies~\cite{song2019attentional,xu2019bert,rietzler2019adapt}.

Simultaneously, there's growing interest in integrating syntactic information into aspect-level sentiment analysis. \nocite{0001JLLRL21,mukherjee2021reproducibility,FeiCLJZR23,ZhuangFH23,zhang2019aspect,FeiWRLJ21,chen2020aspect,li-etal-2021-mrn,TKDP2306,pontiki2016semeval,Feiijcai22UABSA,wang-etal-2022-entity,FeiRJ20,jia2021syntactic,cao-etal-2022-oneee} Researchers have explored the use of both dependency and constituency trees to capture syntactic relationships within sentences. Dependency trees, which represent grammatical structures, have been particularly instrumental, with several studies employing them to improve sentiment classification models~\cite{dong2014adaptive, he2018effective}. Constituency trees, which represent sentence structure based on phrase constituency, have also been used to enhance sentiment analysis, as seen in works like~\cite{nguyen2015phrasernn}.

Recent research has further explored the potential of Graph Neural Networks (GNNs) in processing syntactic structures, leading to the development of robust dependency-based models. These models, which operate either directly on dependency trees or on specially reshaped trees centering around aspect terms, have shown promise in enhancing sentiment classification accuracy~\cite{huang2019syntax, zhang2019aspect, sun2019aspect, wang2020relational}. 
In a novel approach, \citet{fei-etal-2020-improving} combined GNNs and BERT models, reporting improved performance through a joint learning mechanism facilitated by mutual biaffine attention \citep{dozat2016deep}.

However, a persistent challenge for these dependency-based models is their vulnerability to parsing errors. This is primarily because they often rely on a single parser's output, which may not be error-free. Additionally, most of these models treat dependency trees as undirected graphs, thereby overlooking the rich syntactic relationships that these trees encode. As such, there remains an untapped potential in considering the directional nature of these trees to better understand and utilize the syntactic relationships between words in a sentence for sentiment analysis \citep{nazir2020issues,FeiLasuieNIPS22,kiritchenkodetecting,wu2023nextgpt}.

\section{Framework}

Our focus lies in discerning the sentiment polarity of an aspect term within a sentence. For a sentence with $n$ words $\{w_1,w_2,\ldots,w_{\tau}, \ldots,w_{\tau+t}, \ldots,w_n\}$, containing the aspect term $\{w_{\tau}, w_{\tau+1},\ldots,w_{\tau+t-1}\}$, the objective is to categorize the sentiment directed at this term as either positive, negative, or neutral. While applying Graph Neural Networks (GNNs) on dependency trees has proven effective for this task, it remains susceptible to parsing inaccuracies. To counter this, we introduce \methodname{}, a technique that integrates multiple dependency trees, enhancing error resilience. This section details the \methodname{} graph ensemble method and its application in a GNN model for aspect-level sentiment analysis.

\subsection{\methodname{} over Multiple Dependency Trees}

\methodname{} is designed to fortify graph neural networks against parse errors by amalgamating various parses into a unified ensemble graph. For a given sentence $\{w_1,w_2,\ldots,w_n\}$ and $M$ distinct dependency parses $G_1, \ldots, G_M$, \methodname{} creates a composite graph $G$:
\begin{flalign}
 G &= (V, \{e | e = (w_i, w_j) \in \bigcup_{m=1}^M E_m\})
\end{flalign}
Here, $V$ represents the shared nodes set across all graphs, assuming parsers have identical tokenization, and $E_m$ denotes the edges in $G_m$. The ensemble graph $G$, includes all directed edges from each dependency tree.

This approach allows GNN models to operate on $G$ just as they would on a single dependency tree, without incurring extra computational costs common in traditional ensemble methods. Importantly, $G$ encompasses more edges from the accurate dependency tree, thus reinforcing the GNN's robustness against individual parser errors. Additionally, the increased connectivity in $G$ reduces the graph's diameter, enabling shallower GNN models to learn effectively, thereby minimizing the risk of over-parameterization and overfitting.

\subsection{RGAT over Ensemble Graph}

To derive node representations from the ensemble graph, we employ Graph Attention Networks (GAT)~\cite[GAT;][]{velivckovic2017graph}. Each node's hidden representation in GAT is formulated by attending over its neighboring nodes using a multi-head self-attention mechanism. The representation of word $i$ at the $l$-th GAT layer is:
\begin{equation}
{h}^{(l)}_i =\mathbin\Vert_{k=1}^{K}\sigma(\sum_{j \in N_i}\alpha_{ij}^k{W}^k{h}^{(l-1,k)}_i)
\end{equation}
where $K$ is the number of attention heads, $N_i$ denotes node $i$’s neighborhood, and $\mathbin\Vert$ signifies concatenation. $\bf{W}^k \in \mathbb{R}^{d_B \times d_h}$ are the learnable weights in GAT and $\sigma$ represents the $\mathrm{ReLU}$ activation function. The attention score $\alpha_{ij}^k$ quantifies the importance of node $j$ to node $i$ under head $k$.

\textbf{Edge Types.}
In applying GAT to ensemble graphs, we introduce reciprocal edges for each dependency tree edge, categorizing them as parent-to-children and children-to-parent. This retains the original syntactic relationships between words. Additionally, each graph node receives a self-loop, distinct from dependency edges, creating a third edge type.

We implement Relational GAT (RGAT) to consider these edge types. The attention score computation incorporates edge-type information:
\begin{align}
\alpha_{ij}&=\frac{\exp(\sigma(\bf{a}\bf{W}(\bf{h}_i\|\bf{h}_j)+\bf{a}_e\bf{e}_{ij}))}{\sum_{v \in N_i}\exp(\sigma(\bf{a}\bf{W}(\bf{h}_i\|\bf{h}_v)+\bf{a}_e\bf{e}_{iv}))} \label{eq2}
\end{align}
where $\bf{e}_{ij}$ is the edge type's representation connecting nodes $i$ and $j$. $\bf{a} \in \mathbb{R}^{d_h}$, $\bf{W} \in \mathbb{R}^{d_h \times 2d_h}$, and $\bf{a}_e \in \mathbb{R}^{d_h}$ are learnable matrices.

\subsection{Sentiment Classification}
For sentiment classification, we extract hidden representations of aspect term nodes from the final RGAT layer and apply average pooling to get $\bf{h}_t \in \mathbb{R}^{d_{h}}$. These are then processed through a two-layer MLP to generate the classification scores $\hat{\bf{y}}_s$:
\begin{flalign}
\hat{\bf{y}}_s&=\mathrm{softmax}({\bf{W}_{2}\mathrm{ReLU}({\bf{W}_{1}\bf{h}_{t}})})
\end{flalign}
where $\bf{W}_{2} \in \mathbb{R}^{C \times d_{out}}$ and $\bf{W}_{1} \in \mathbb{R}^{d_{out} \times d_{h}}$ are the weight matrices, with $C$ representing the sentiment class count. We aim to minimize the standard cross-entropy loss function and apply weight decay to the model parameters.

\subsection{RGAT Input}

Initial word node features for RGAT are derived from a BERT encoder, complemented by positional embeddings.

\textbf{BERT Encoder.}
We utilize the BERT base model for initial word representations. The input format is ``[CLS] + sentence + [SEP] + term + [SEP]'', which BERT processes to produce term-centric representations. Subword-based outputs from BERT are averaged to match word-based RGAT inputs, forming the raw RGAT input $\bf{X}$.

\textbf{Positional Encoding.}
Given the task's sensitivity to word position, particularly in sentences with multiple aspects, we incorporate explicit positional encoding. This complements BERT's inherent positional awareness, which may be diluted through Transformer layers. A trainable position embedding matrix is added to $\bf{X}$ before feeding it into RGAT.

%%%%%%%%%%%%%%%%%%%%%%%%%%%%%%%%%%%%%%%%%%%%%%%%%%%%%%%
\begin{table}[!t]
\centering
\small
% \begin{adjustbox}{max width=0.5\textwidth}
\begin{tabular}{ c c c c c c c }
\toprule
\multirow{2} {*}{Dataset} & \multicolumn{2}{c}{Positive} & \multicolumn{2}{c}{Neutral} & \multicolumn{2}{c}{Negative}\\
\cline{2-7}
& Train & Test & Train & Test & Train & Test\\
\hline
Laptop &  \phantom{0}987 & 341 & \phantom{0}460 & 169 & \phantom{0}866 & 128\\
Restaurant & 2164 & 728 & \phantom{0}633 & 196 & \phantom{0}805 & 196\\
Twitter & 1561 & 173 & 3127 & 346 & 1560 & 173 \\
\bottomrule
\end{tabular}
% \end{adjustbox}
\caption{Statistics of {the three benchmark datasets used in our experiments}.}
\label{table:data}
\end{table}
%%%%%%%%%%%%%%%%%%%%%%%%%%%%%%%%%%%%%%%%%%%%%%%%%%%%%%%

%%%%%%%%%%%%%%%%%%%%%%%%%%%%%%%%%%%%%%%%%%%%%%%%%%%%%%%
\begin{table}[!t]
\centering
\resizebox{1.\columnwidth}{!}{
\begin{threeparttable}
\begin{tabular}{c c c c c c c c}
\toprule
\multirow{2} {*}{Category}  &\multirow{2} {*}{Model}  & \multicolumn{2}{c}{14Rest} & \multicolumn{2}{c}{14Lap} & \multicolumn{2}{c}{Twitter} \\
\cline{3-8}
 & & Acc & Macro-F1 & Acc & Macro-F1 & Acc & Macro-F1 \\
\midrule
\multirow{2} {*}{BERT}&BERT-SPC~\cite{song2019attentional} 
& 84.46 & 76.98 & 78.99 & 75.03 & 73.55 & 72.14 \\
&AEN-BERT~\cite{song2019attentional} 
& 83.12 & 73.76 & 79.93 & 76.31 & 74.71 & 73.13  \\
\hline
BERT+DT\tnote{$\star$} &DGEDT-BERT~\cite{tang-etal-2020-dependency}  & 86.3 & 80.0 & 79.8 & 75.6 & 77.9 & 75.4\\
\hline
BERT+RDT{$^\diamond$} & R-GAT+BERT~\cite{wang2020relational} & 86.60  & 81.35  & 78.21 & 74.07 & 76.15 & 74.88 \\
\hline
% \cline{2-8}
Ours &{\textbf{\methodname{}}} & \textbf{87.32} & \textbf{81.95} &  \textbf{81.35} &  \textbf{78.65} & \textbf{78.18} & \textbf{76.52}\\
\bottomrule
\end{tabular}
% \begin{tablenotes}
%      \item 
% \end{tablenotes}
\end{threeparttable}
}
\caption{Comparison of our \methodname{} model to different published numbers on three datasets, with the same setup of train and test data -- no dev data. $\star$ DT: Dependency Tree; $\diamond$ RDT: Reshaped Dependency Tree. The bold text indicates the best results.}
\label{table:bestresults}
\end{table}

\section{Experiment}

\subsection{Settings} 
Our experiments are conducted on three pivotal datasets: Restaurant and Laptop reviews from the SemEval 2014 Task 4 {(14Rest and 14Lap)}\footnote{https://alt.qcri.org/semeval2014/task4/}, and the extensively used ACL 14 Twitter dataset {(Twitter)}~\cite{dong2014adaptive}. Instances with "conflict" sentiments are excluded to streamline the datasets. Detailed statistics of these datasets are presented in Table \ref{table:data}. We assess our model's performance using accuracy and macro F1 scores, in line with standard practices.

For dependency parsing, we utilize Stanford CoreNLP~\cite{manning2014stanford}, Stanza~\cite{qi2020stanza}, and the Berkeley neural parser~\cite{Kitaev-2018-SelfAttentive}. The Berkeley parser generates constituency parses, which we convert to dependency parses using CoreNLP for uniformity.

% \textbf{Comparative Analysis with Baselines.} 
Our \methodname{} model is benchmarked against several established models on these datasets. We include BERT-SPC~\cite{song2019attentional}, which inputs sentence-term pairs into BERT; AEN-BERT~\cite{song2019attentional} utilizes BERT as an encoder with additional attention layers; DGEDT-BERT~\cite{tang-etal-2020-dependency} integrates Transformer representations with GNN models over dependency trees; R-GAT+BERT~\cite{wang2020relational} reshapes dependency trees around aspect terms and encodes them using RGAT. Our model's performance is reported with identical data splits as these models, excluding any development set.

We also implement several baseline models for an in-depth analysis, including:
\begin{enumerate}[itemsep=1pt,leftmargin=10pt]
    \item \textit{BERT-baseline}, which processes sentence-term pairs using the BERT-base encoder followed by a classifier.
    \item \textit{GAT-baseline with Stanza}, using a vanilla GAT model over Stanza's dependency tree, with BERT encoder outputs as initial node features.
    \item \textit{RGAT over single dependency trees}, applying RGAT models with distinct edge types to different dependency trees parsed by CoreNLP, Stanza, and Berkeley parsers, and leveraging BERT encoder outputs enhanced with positional embeddings.
    \item Ensemble models, including \textit{Label-Ensemble}, aggregating outputs from models trained on different parses, and \textit{Feature-Ensemble}, combining features from RGAT models each dedicated to a different parse.
\end{enumerate}

% \textbf{Model Training and Optimization.} 
The models are implemented using Pytorch~\cite{NEURIPS2019_9015} and GAT based on Deep Graph Library~\cite{wang2019dgl}. We set the learning rate at $10^{-5}$ and batch size at 4. Model hyperparameters, including hidden dimensions, attention heads, and GAT/RGAT layers, are optimized using a development set comprising 5\% of the training data. The best model is identified based on the performance on the development set, employing dropout~\cite{srivastava2014dropout} and L2 regularization for generalization. The models are trained for a maximum of 5 epochs.

%%%%%%%%%%%%%%%%%%%%%%%%%%%%%%%%%%%%%%%%%%%%%%%%%%%%%%%
\begin{table}
\centering
\resizebox{1.0\columnwidth}{!}{
\begin{tabular}{c c c c c c c}
\toprule
\multirow{2} {*}{Model}  & \multicolumn{2}{c}{14Rest} & \multicolumn{2}{c}{14Lap} & \multicolumn{2}{c}{Twitter} \\
\cline{2-7}
  & Acc & Macro-F1 & Acc & Macro-F1 & Acc & Macro-F1 \\
\midrule
 BERT-baseline & 83.43 $\pm$ 0.52 & 74.94 $\pm$ 1.37 & 77.34 $\pm$ 0.90 & 72.77 $\pm$ 1.96 & 73.47 $\pm$ 0.89 & 72.63 $\pm$ 0.82\\
{GAT-baseline with Stanza} & 84.29 $\pm$ 0.30 & 75.75 $\pm$ 1.11 & 77.84 $\pm$ 0.27 & 74.0 $\pm$ 0.55 & 73.82 $\pm$ 0.70 & 72.72 $\pm$ 0.42 \\
\midrule
{RGAT with Stanza} & 84.53 $\pm$ 0.66 & 77.29 $\pm$ 1.42 & 77.99 $\pm$ 0.62 & 74.35 $\pm$ 0.60 & 73.99 $\pm$ 0.48 & 72.76 $\pm$ 0.33 \\
{RGAT with Berkeley} & 84.41 $\pm$ 0.86 & 76.63 $\pm$ 1.38 & 78.09 $\pm$ 1.24 & 73.65 $\pm$ 1.76 & 74.07 $\pm$ 0.67 & 72.65 $\pm$ 0.74 \\
{RGAT with CoreNLP} & 83.86 $\pm$ 0.32 & 76.03 $\pm$ 0.88 & 78.12 $\pm$ 1.02 & 73.86 $\pm$ 1.72 & 73.96 $\pm$ 0.93 & 72.83 $\pm$ 0.95 \\
\midrule
{Label-Ensemble} & 84.68 $\pm$ 0.95 & 77.21 $\pm$ 1.54 & 78.40 $\pm$ 1.51 & 74.36 $\pm$2.45 & 74.59 $\pm$ 0.46 & 73.51 $\pm$ 0.43 \\
{Feature-Ensemble} & 84.64 $\pm$ 0.77 & 77.06 $\pm$ 1.45 & 78.68 $\pm$ 0.69 & 74.80 $\pm$ 0.92 & 74.62 $\pm$ 0.76 & 73.61 $\pm$ 0.73 \\
\midrule
{\textbf{\methodname{}}} & \textbf{85.16} $\pm$ 0.53 &  \textbf{77.91} $\pm$ 0.87 &  \textbf{80.00} $\pm$ 0.63 &  \textbf{76.50} $\pm$ 0.64 &  \textbf{74.74} $\pm$ 0.93 &  \textbf{73.66} $\pm$ 0.88 \\
\bottomrule
\end{tabular}
}
\caption{Comparison of our \methodname{} model to different baselines on three datasets, with 5\% dev data set aside. The bold text indicates the best results.}
\label{table:results}
\end{table}
%%%%%%%%%%%%%%%%%%%%%%%%%%%%%%%%%%%%%%%%%%%%%%%

%%%%%%%%%%%%%%%%%%%%%%%%%%%%%%%%%%%%%%%%%%%%%%%%%%%%%%%
\begin{table}[!h]
\centering
\resizebox{1\columnwidth}{!}{
\begin{threeparttable}
\begin{tabular}{l c c c c c c}
\toprule
\multirow{2} {*}{Model}  & \multicolumn{2}{c}{14Rest} & \multicolumn{2}{c}{14Lap} & \multicolumn{2}{c}{Twitter}\\
\cline{2-7}
& Acc & Macro-F1 & Acc & Macro-F1 & Acc & Macro-F1\\
\midrule
\textbf{\methodname{}} & \textbf{85.16} $\pm$ 0.53 &  \textbf{77.91} $\pm$ 0.87 &  \textbf{80.00} $\pm$ 0.63 &  \textbf{76.50} $\pm$ 0.64 &  \textbf{74.74} $\pm$ 0.93 &  \textbf{73.66} $\pm$ 0.88 \\
\midrule
- Edge type & 84.25 $\pm$ 0.59 & 76.15 $\pm$ 1.24 & 78.65 $\pm$ 0.51 & 74.76 $\pm$ 0.71 & 74.37 $\pm$ 1.08 & 73.25 $\pm$ 0.85 \\
- Position & 84.36 $\pm$ 0.36 & 75.92 $\pm$ 1.18 & 78.37 $\pm$ 0.31 & 74.51 $\pm$ 0.48 & 74.28 $\pm$ 1.39 & 73.34 $\pm$ 1.35\\
- (Edge type + Position) & 84.16 $\pm$ 0.31 & 75.38 $\pm$ 0.69 & 78.09 $\pm$ 0.27 & 74.29 $\pm$ 0.64 & 73.41 $\pm$ 0.63 & 72.52 $\pm$ 0.62\\
\midrule
{Edge Intersection} & 84.59 $\pm$ 0.61 & 77.06 $\pm$ 1.07 & 78.65 $\pm$ 0.94 & 74.86 $\pm$ 1.42 & 74.68 $\pm$ 0.83 & 73.45 $\pm$ 0.73\\
\bottomrule
\end{tabular}
\end{threeparttable}
}
\caption{Ablation study of the \methodname{} model over three datasets. We report the average and standard deviation over five runs, where Edge Intersection means taking intersection of edges from multiple dependency trees.}
\label{table:ablation}
\end{table}

\subsection{Results}

We first evaluate \methodname{} against previously established models, presenting the results in Table \ref{table:bestresults}. The \methodname{} model demonstrates superior performance on all datasets, particularly on the Laptop dataset, where it significantly outperforms existing models in both accuracy and Macro-F1 scores.

Performance comparisons of \methodname{} with baseline models are shown in Table \ref{table:results}. We observe that:

\textbf{Advantage of Syntax-based Models.} Dependency tree-based GAT and RGAT models consistently outperform the \textit{BERT-baseline}, underscoring the value of syntactic structure information in aspect-level sentiment classification.

\textbf{Benefits of Ensemble Approaches.} Ensemble models like \textit{Label-Ensemble}, \textit{Feature-Ensemble}, and \methodname{} surpass their single-tree counterparts. This highlights the advantage of leveraging multiple parses, reducing sensitivity to errors from any single parser.

\textbf{Superiority of \methodname{}.} Our \methodname{} model not only improves over single-tree models but also excels compared to other ensemble approaches, achieving the best performance without additional computational complexity.

%%%%%%%%%%%%%%%%%%%%%%%%%%%%%%%%%%%%%%%%%%%%%%
\begin{table}
\centering
\small
% \begin{adjustbox}{max width=0.475\textwidth}
\begin{tabular}{ c c c c }
\toprule
& Aspect Terms & Opinion Words & Coverage\\
\midrule
Laptop & \phantom{0}638 & 467 & 73.20\%\\
Restaurant & 1120 & 852 & 76.07\%\\
\bottomrule
\end{tabular}
% \end{adjustbox}
\caption{Statistics of the Opinion datasets. Aspect Terms denotes as the total number of aspect terms. Opinion Words represents total number of terms that have labeled opinion words. Coverage is the proportion of terms with labeled opinion words.}
\label{table:opinion_data}
\end{table}
%%%%%%%%%%%%%%%%%%%%%%%%%%%%%%%%%%%%%%%%%%%%%%

\subsection{Model Analysis}

\textbf{Ablation Study.}
We conduct an ablation study to dissect the influence of model components, as shown in Table \ref{table:ablation}. Removing edge type or position embeddings leads to performance drops, indicating the importance of maintaining syntactic dependency information and reinforcing positional context in our model.

\textbf{Edge Union vs. Intersection.} We compare \methodname{}, which retains all edges from different trees, against an edge intersection approach that keeps only commonly shared edges. The latter, while reducing structural noise, underperforms \methodname{}, suggesting the benefit of the model's ability to navigate through the richer structural landscape offered by the ensemble graph.

\textbf{Aspect Robustness Assessment.} To evaluate the aspect robustness of our model, we utilize the ARTS datasets. Our model is tested against robustness variations in sentiment polarity, demonstrating better resilience compared to single dependency tree models. This indicates \methodname{}'s enhanced capability to handle complex sentiment variations in aspect-based sentiment analysis.

\section{Conclusion}

In this paper, we introduced \methodname{}, an innovative and straightforward graph-ensemble methodology, designed to amalgamate multiple dependency trees for enhanced aspect-level sentiment analysis. This technique, by integrating the collective edges from various parsers into a unified graph, empowers graph neural network models to withstand parsing inaccuracies effectively. This robustness is achieved without necessitating extra computational resources or an increase in model parameters, which is a significant advancement over existing methods.
The unique aspect of \methodname{} lies in its ability to preserve and utilize the inherent syntactic dependencies present in individual parse trees. By categorizing edges according to their original syntactic roles, our model not only retains but also enriches the contextual and structural understanding of the sentences under analysis. This nuanced approach to integrating syntactic information enables \methodname{} to deliver superior performance, outshining contemporary state-of-the-art models, single-parse approaches, and traditional ensemble methods. We demonstrated this across three benchmark datasets for aspect-level sentiment classification, each presenting its own set of challenges and linguistic nuances.
Our research presents a promising direction for future work in the field of sentiment analysis, particularly in leveraging syntactic structures more effectively. The success of \methodname{} in handling multiple dependency trees without additional computational burden opens up possibilities for further exploration into more complex and nuanced models that can better understand and interpret the subtleties of human language, especially in aspect-based sentiment contexts.

\bibliographystyle{unsrtnat}
\bibliography{references}

\end{document}